\documentclass[10pt,twocolumn,letterpaper]{article}

\usepackage[pagenumbers]{iccv} 

\definecolor{iccvblue}{rgb}{0.21,0.49,0.74}
\usepackage[pagebackref,breaklinks,colorlinks,allcolors=iccvblue]{hyperref}

\usepackage{graphicx}	
\usepackage{amsmath}	
\usepackage{amssymb}	
\usepackage{booktabs}
\usepackage{times}
\usepackage{microtype}
\usepackage{epsfig}
\usepackage{caption}
\usepackage{float}
\usepackage{placeins}
\usepackage{color, colortbl}
\usepackage{stfloats}
\usepackage{enumitem}
\usepackage{tabularx}
\usepackage{xstring}
\usepackage{multirow}
\usepackage{xspace}
\usepackage{url}
\usepackage{subcaption}
\usepackage[hang,flushmargin]{footmisc}

\def\paperID{2398} 
\def\confName{ICCV}
\def\confYear{2025}

\title{Visual Acoustic Fields}

\author{
    Yuelei Li\textsuperscript{1}\ \hspace{1em} 
    Hyunjin Kim\textsuperscript{1}\ \hspace{1em}    
    Fangneng Zhan\textsuperscript{2,3} \ \hspace{1em} 
    Ri-Zhao Qiu\textsuperscript{1}\ \hspace{1em} 
    Mazeyu Ji\textsuperscript{1}\ \hspace{1em}  
    Xiaojun Shan\textsuperscript{1}\\  
    Xueyan Zou\textsuperscript{1}\ \hspace{1em}  
    Paul Liang\textsuperscript{3}\ \hspace{1em} 
    Hanspeter Pfister\textsuperscript{2}\ \hspace{1em} 
    Xiaolong Wang\textsuperscript{1}\\ 
    \textsuperscript{1}UC San Diego \ \ 
    \textsuperscript{2} Harvard University \ \
    \textsuperscript{3}MIT
}

\begin{document}

\twocolumn[{
    \renewcommand\twocolumn[1][]{#1}
    \maketitle
    \begin{center}
        \centering
        \captionsetup{type=figure}
        \includegraphics[width=\textwidth]{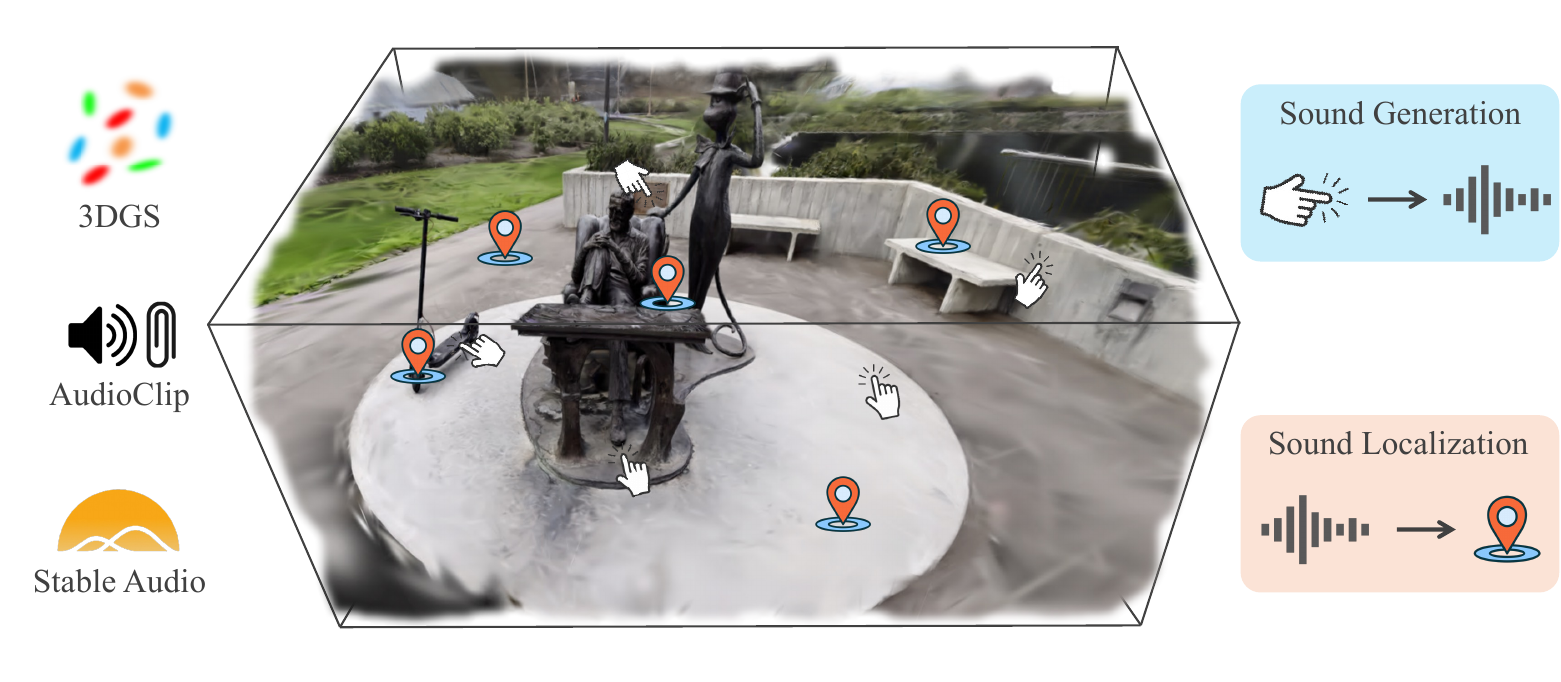}
        \vspace{0.05in}
        \captionof{figure}{
        Overview of Visual Acoustic Fields, a novel framework for integrating visual and auditory signals within a 3D scene. Our approach leverages 3D Gaussian Splatting (3DGS) to represent the scene and associates it with impact sounds. The framework enables two key tasks: vision-conditioned sound generation, where impact sounds are synthesized based on impact location, and sound localization, where the model identifies the source of a given sound within the 3D environment.
        }
        \label{fig_teaser}
        
    \end{center}
}]

\begin{abstract}
Objects produce different sounds when hit, and humans can intuitively infer how an object might sound based on its appearance and material properties. Inspired by this intuition, we propose Visual Acoustic Fields, a framework that bridges hitting sounds and visual signals within a 3D space using 3D Gaussian Splatting (3DGS). Our approach features two key modules: sound generation and sound localization. The sound generation module leverages a conditional diffusion model, which takes multiscale features rendered from a feature-augmented 3DGS to generate realistic hitting sounds. Meanwhile, the sound localization module enables querying the 3D scene, represented by the feature-augmented 3DGS, to localize hitting positions based on the sound sources. 
To support this framework, we introduce a novel pipeline for collecting scene-level visual-sound sample pairs, achieving alignment between captured images, impact locations, and corresponding sounds.
To the best of our knowledge, this is the first dataset to connect visual and acoustic signals in a 3D context. 
Extensive experiments on our dataset demonstrate the effectiveness of Visual Acoustic Fields in generating plausible impact sounds and accurately localizing impact sources. Our project page is at \href{https://yuelei0428.github.io/projects/Visual-Acoustic-Fields/}{https://yuelei0428.github.io/projects/Visual-Acoustic-Fields/}
\end{abstract}    
\section{Introduction}
\label{sec:intro}
Our lives are filled with objects that produce distinct sounds. For example, striking a ceramic cup and a wooden table in a living room scene produces completely different sounds. Studying the cross-modal connection between vision and impact sounds is important because it represents one of the most fundamental ways we learn about the physical world: infants explore their surroundings by simultaneously observing and interacting with objects, and studies have shown that this process helps them develop an intuitive understanding of physics \cite{originofinquiry, isola2015learningvisualgroupscooccurrences,baillargeon2002acquisition}. Such an understanding can enhance applications that require reasoning about the physical properties of interactive objects, such as robotics \cite{conceptfusion,hong2024multiply,huang2023audio}, virtual reality \cite{su2023physics,chen2024action2sound}, and content creation \cite{xing2024seeing,zhang2024foleycrafter}.

Existing cross-modal datasets that connect vision and sound typically pair a whole 2D image or video depicting a large scene with a soundtrack. These datasets encompass a diverse range of sounds, including impact sounds \cite{owens2016visually,gao2023objectfolder,conceptfusion}, speech \cite{wang2020mead,chung2017lip,afouras2018deep}, and background music \cite{li2021ai,zhu2022quantized}. However, in all these datasets, the paired audio represents a holistic soundscape that describes the entire scene rather than isolating the specific source producing the sound. As a result, such datasets do not provide physical information about the precise sound source within the scene. Furthermore, they do not capture the relationship between auditory and visual signals in 3D.


In this work, we aim to collect spatially aligned visual-sound pairs in 3D space. This is a challenging task because we must determine the location of each local visual signal in the scene (e.g., a subpart of an object) within the entire scene, and this centimeter-level alignment of visual supervision and audio supervision is hard to achieve.
Since no existing sound dataset is available for 3D scenes, we refer to visual-tactile datasets, which commonly involve a similar step of localizing signals within a scene. Early work relies on robotic arms capturing signals in controlled settings, where the gripper's location is determined via forward kinematics \cite{calandra2017feeling,calandra2018more,kerr2022self,li2019connecting}.
\cite{tactileaugmentedradiancefields} moves a step forward to enable signal localization at the scene level, but it requires a specialized device and a complex camera calibration step.

To solve the above problems, we propose an easy-to-use pipeline for localizing data at the scene level without requiring any new devices or calibration. Our approach relies solely on a smartphone to capture images and record sounds. First, we collect a set of multiview images of the scene. Next, we capture a set of labeled images, each corresponding to a location where an impact occurs, along with the associated sound. We use structure-from-motion \cite{schoenberger2016mvs, schoenberger2016sfm} to estimate the camera poses for all images. The multiview images are then used to reconstruct the scene using 3DGS~\cite{kerbl20233d}, and the camera poses of the labeled images help determine the impact locations within the reconstructed scene.

With the collected dataset, we propose \textbf{Visual Acoustic Fields} to bridge visual and auditory signals in 3D scenes (see Fig. \ref{fig_teaser}). 
Based on 3DGS~\cite{kerbl20233d}, the Visual Acoustic Fields learn a set of 3D Gaussians augmented with AudioCLIP~\cite{audioclip} features, which are supervised by the AudioCLIP embeddings of the collected multiview images.
The AudioCLIP associates the visual and auditory signals, allowing to perform two tasks including (1) vision-conditioned sound generation and (2) sound localization.
For sound generation from impact points, we infer the AudioCLIP feature of the impact point, which is further mapped to sound with an audio diffusion model. Specifically, we modify and finetune a pre-trained Stable Audio model \cite{stableaudioopen} for the mapping to take advantage of its generalization ability. For sound localization, we adopt contrastive pretraining~\cite{clip, audioclip} to finetune AudioCLIP on our collected dataset, which is used to encode the query sound. Then, a region or object can be localized by measuring the relevancy between the sound and visual encoding.

We conduct experiments on our collected dataset to evaluate our sound generation and localization models. Our experiments indicate that:
\begin{itemize}
  \item The impact locations of collected visual-sound pairs can be localized in 3D space by learning a radiance field with synchronized camera poses.
  \item Predicted impact sounds in our Visual Acoustic Fields accurately align with the corresponding impact locations.
  \item Impact regions or objects can be precisely retrieved with sound using our Visual Acoustic Fields.
\end{itemize}

\section{Related Work}
\label{sec:related}

\subsection{Visual-Sound Datasets.}
Most existing cross-modal datasets that connect vision and sound focus on providing general background soundtracks for given images (e.g., a cheerful soundtrack for images of a sunny outdoor scene, etc.) \cite{audioset, UrbanSound, piczak2015esc}. These general soundtrack datasets are easier to collect at scale from the Internet, whereas high-quality hitting sounds are not as readily available online. To date, only two datasets have specifically focused on hitting sounds: ObjectFolder \cite{ gao2021objectfolder, gao2022objectfolder, gao2023objectfolder} and The Greatest Hits Dataset \cite{owens2016visually}. ObjectFolder includes representations of individual neural objects in the form of implicit neural fields with simulated multisensory data. The Greatest Hits Dataset features videos of objects struck by a drumstick without the label for hitting position. Our dataset differs from these in two significant aspects: 1) it operates at the scene level rather than the object level, and 2) the collected visual-sound pairs are spatially registered within a 3D scene represented by 3DGS.


\subsection{Predicting Sound from Visual Inputs}

Predicting sound from visual inputs is a fundamental problem in cross-modal learning with applications in robotics, virtual reality, and content generation. 
Most methods focus on a holistic generation that yields a single soundtrack for the whole scene
\cite{sheffer2023iheartruecolors,luo2023difffoleysynchronizedvideotoaudiosynthesis,v2amapper}.
For instance, Owens et al.~\cite{owens2016visually} introduced a self-supervised model for impact sound generation, while Zhou et al.~\cite{zhou2018visual} learned direct mappings from video frames to sound representations. More recently, Sheffer and Adi~\cite{sheffer2023hear} developed \textit{im2wav}, which conditions sound synthesis on images but lacks spatial awareness.

Beyond scene-wide generation, recent approaches explore object-specific impact sound prediction. Gan et al.~\cite{gan2020foley} introduce a system capable of synthesizing plausible music for silent video clips depicting people playing musical instruments.
Multimodal and physics-informed approaches further improve accuracy by integrating visual, auditory, and tactile data.
Su et al.~\cite{su2023physics} leveraged physics-driven simulation for realistic sound modeling, and Zhou et al.~\cite{zhou2022audio} incorporated material-aware conditioning into audio-visual segmentation.
Gao et al.~\cite{gao2021objectfolder,gao2022objectfolder} developed multisensory datasets, \textit{ObjectFolder} and \textit{ObjectFolder 2.0}, enabling cross-modal learning for object interactions. 
However, the above works fall short in sound generation in 2D space or virtual scenarios.
In contrast, our method allows interaction with the real 3D environment by navigating in the learned Visual Acoustic Fields, enabling fine-grained impact sound synthesis at arbitrary 3D locations.


\subsection{Sound Localization}
Sound localization refers to the task of identifying the source of a sound within a scene. 
Many works leverage the natural synchronization between visual and auditory information to achieve localization. Arandjelovic and Zisserman~\cite{arandjelovic2018objects} introduced a \textit{self-supervised} approach to associate \textit{spatial regions} in video frames with corresponding sounds, providing \textit{pixel-level} audio localization. Similarly, Zhao et al.~\cite{zhao2019sound} proposed to localize sound sources by leveraging temporal coherence in video sequences, achieving \textit{patch-level} localization.
Besides localization, recent works have also explored \textit{audio-visual segmentation (AVS)}.
Zhou et al.~\cite{zhou2022audio} proposed a transformer-based model that applies \textit{pixel-wise attention} to capture detailed audio-visual correspondences.
To include explicit object-level alignment,
Huang et al.~\cite{huang2023discovering}  defines \textit{audio queries} to explicitly associate sound features with individual objects and improves sound localization accuracy.

While most audio-visual learning methods operate on 2D video frames, their applicability remains limited in 3D. 
Jatavallabhula et al.~\cite{conceptfusion} proposed ConceptFusion, which integrates audio and visual signals within a 3D representation, enabling object localization in 3D scenes. However, this method does not focus on hitting sounds; neither its sound localization model nor its dataset has been open-sourced.
In contrast, we will open-source both our model and dataset.

\begin{figure}[ht]
    \includegraphics[width=1.0\linewidth]{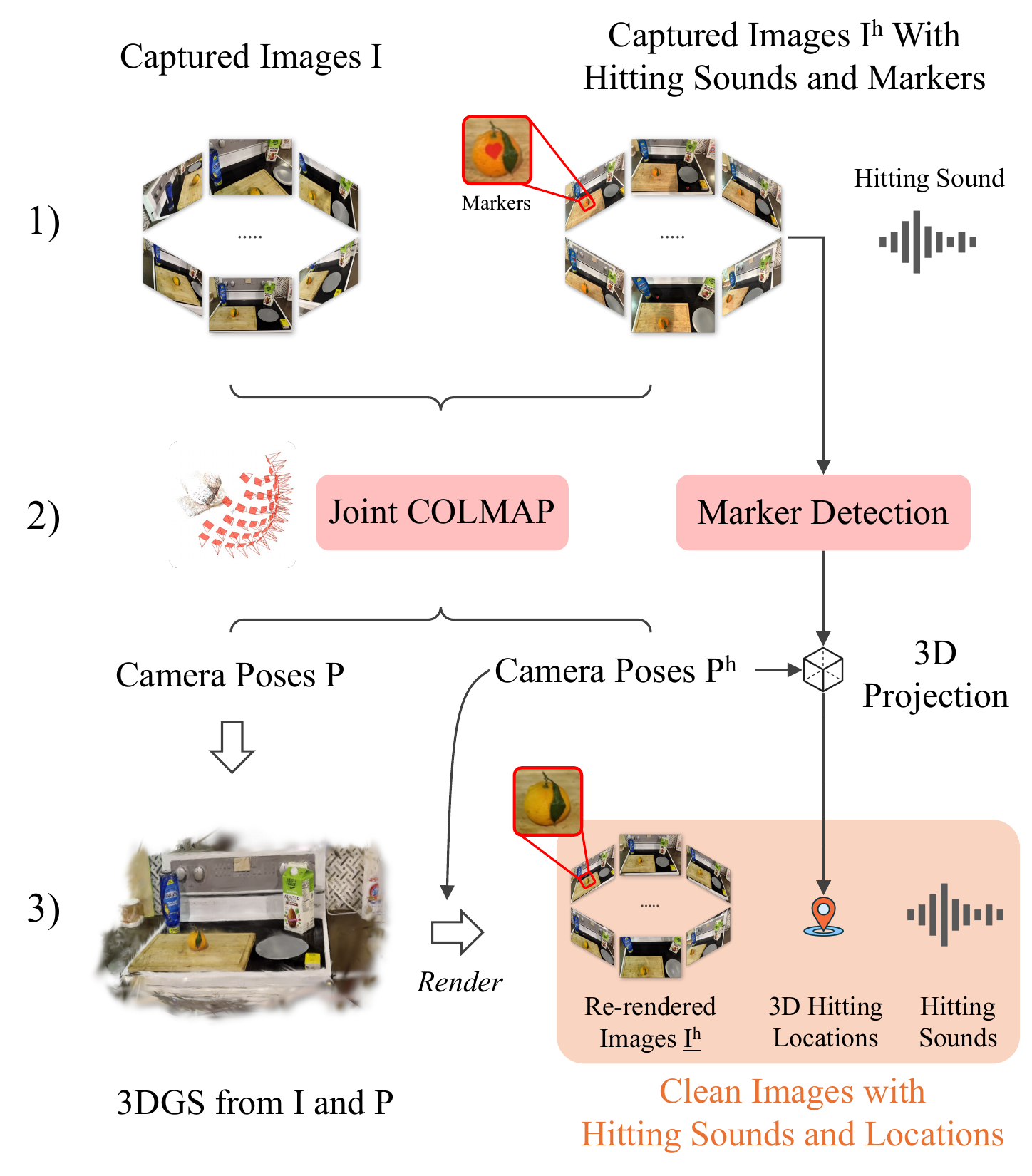}
    \caption{
    \textbf{Pipeline for data collection.}
    A novel re-rendering strategy is proposed to enable accurate annotation of impact sounds and their locations without introducing artifacts. 
    1) We capture two sets of multiview images including \( I \) of the scene and \( I^h \) marked with visible hitting markers and synchronized with corresponding hitting sounds.
    2) Using \textit{Structure-from-Motion (SfM)}, we jointly estimate camera poses of \( I \) and \( I^h \), as denoted by \( P \) and \( P^h \), respectively. The impact locations can be obtained by detecting the markers with OWL-v2 \cite{minderer2023scaling}, which are further projected to 3D location with known camera poses \( P^h \) and depth map.
    3) A 3DGS can be trained with multiview images \( I \) and camera poses \( P \). The images with impact locations are re-rendered without markers from the 3DGS with camera poses \( P^h \), yielding clean images $\underline{I^h}$ with paired hitting sounds and their hitting positions.
    }
    \label{fig_data_collection}
\end{figure}

\section{Method}
\label{sec:method}

\subsection{Data Collection Pipeline.}

To train our Visual Acoustic Fields, we require multiview images annotated with both impact sounds and their corresponding locations. However, collecting such a dataset in the real world is a challenging task.
While hitting sounds can be recorded using appropriate audio equipment, accurately identifying the impact locations in the captured multiview images remains a significant challenge. A straightforward solution is to use markers to label these locations; however, the presence of markers introduces artifacts into the images, which can interfere with model training.
To address this issue, we propose a novel re-rendering strategy for data collection, as illustrated in Fig. \ref{fig_data_collection}.

\textbf{1) Image Capturing and Sound Collection.}
For each scene, we capture two sets of images: \( I = \{i_n\}_{n=1}^N \) and \( I^h = \{i^h_m\}_{m=1}^M \) with size $H \times W$. The set \( I \) consists of multiview images of the scene, collected by moving through the environment while recording a video to ensure dense 3D coverage. The set \( I^h \) comprises images where hitting is performed, with markers indicating the corresponding hitting positions. These markers can be small stickers, laser-projected patterns, or other convenient visual indicators.

For the markers in each image in $I^h$, we record a corresponding hitting sound. A metallic coffee stick is used to strike all the marker locations, ensuring consistency across all data. To remove background noise, we apply spectral gating, which estimates the noise profile using short-time Fourier transform (STFT), subtracts it from the signal, and reconstructs the denoised audio via inverse STFT. We then crop or pad the recording to 0.5 seconds to include the clean hitting sound. Since the force applied when hitting different locations may vary slightly during data collection, resulting in differences in sound amplitude, we apply Root Mean Square (RMS) normalization with a scale of 0.01 to standardize the inputs for our models.

\textbf{2) Joint COLMAP and Hitting Localization} To estimate camera poses, we use COLMAP \cite{schoenberger2016sfm, schoenberger2016mvs}, obtaining pose sets \( P = \{p_n\}_{n=1}^N \) for \( I \) and \( P^h = \{p^h_m\}_{m=1}^M \) for \( I^h \). However, because COLMAP assigns a random coordinate origin during each execution, separately estimating the poses of \( I \) and \( I^h \) results in inconsistent camera coordinate systems between \( P \) and \( P^h \). To ensure alignment, we perform joint pose estimation, processing both sets of images together in a single COLMAP run. Notably, the markers in \( I^h \) are small enough to have a negligible impact on COLMAP’s accuracy in practice.

To locate the collected sound within the 3DGS, we first apply an object detection network OWL-v2 \cite{minderer2023scaling} on the $\{i^h_n\}_{n=1}^N$ to obtain the pixel locations $\{(x^h_n, y^h_n)\}_{n=1}^N$ of the markers.
With $\{(x^h_n, y^h_n)\}_{n=1}^N$ and $\{i^h_n\}_{n=1}^N$, we can follow the standard pinhole camera model to locate the hitting point in the camera coordinate frame:
\begin{equation}
    (i_n, j_n,k_n) = \left( \frac{(x^h_n - c_x) \cdot Z_c}{f_x}, \frac{(y^h_n - c_y) \cdot Z_c}{f_y}, d_n \right)
\end{equation}
where $(x^h_n, y^h_n)$ is the detected marker position in image $i^h_n$, $d_n$ is the depth value at $(x^h_n, y^h_n)$ estimated by the 3DGS, $f_x$ and $f_y$ are the focal lengths in pixels, and $(c_x, c_y)$ is the principal point of the camera.

\textbf{3) Re-rendering.} 
We reconstruct a clean 3D scene without markers from images $I$ and poses $P$ with 3DGS.
Notably, camera poses $P$ and $P^h$ are in the same coordinate systems thanks to our joint camera pose estimation with COLMAP.
Thus, to obtain a clean version (without markers) of $I^h$, 
we can query the 3DGS with camera poses $P^h$ to re-render a new set of images without markers, which we denote as $\underline{I^h}$.
Finally, we obtain a dataset with paired multiview images, hitting locations, and hitting sounds.

\begin{figure*}[t]
    \includegraphics[width=1.0\linewidth]{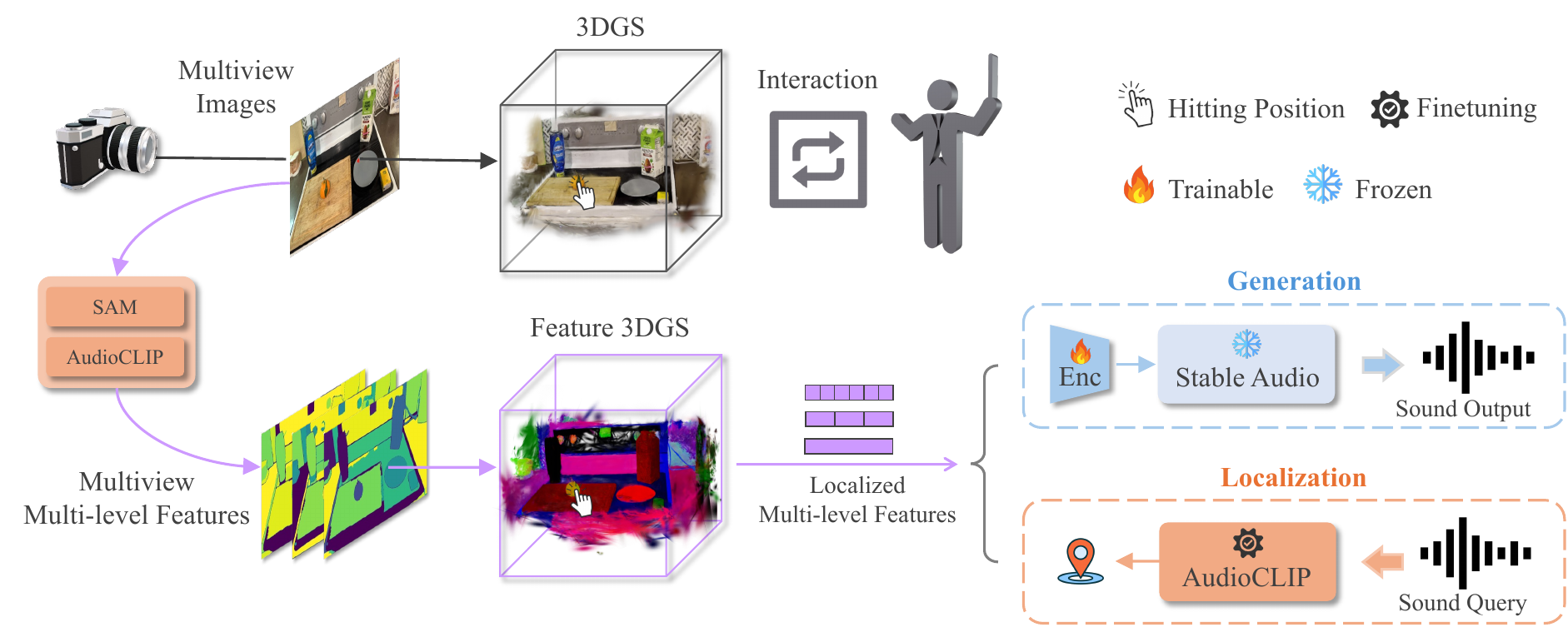}
    \caption{
    Overview of the Visual Acoustic Fields framework. The model consists of two main components: sound generation and sound localization. Given multiview images, a feature-augmented 3D Gaussian Splatting (feature 3DGS) representation is constructed. For sound generation, localized multi-level features queried from the feature 3DGS are used as conditions to fine-tune a pretrained Stable Audio diffusion model to synthesize impact sounds. For sound localization, a fine-tuned AudioCLIP encoder maps input audio queries to the feature 3DGS, allowing the model to localize the corresponding impact location by computing feature similarity. Trainable, frozen, and fine-tuned components are indicated in the diagram.
}
    \label{fig_framework}
\end{figure*}

\subsection{Predicting Sound from Visual Signals}
With the collected dataset, we can train a model that can map visual inputs and hitting positions to the corresponding hitting sounds.
However, accurately estimating the impact sounds requires identifying the specific object or region producing the impact, associating it with relevant acoustic features, and synthesizing realistic sound variations. To achieve this, we incorporate three key components: Segment Anything Model (SAM) for object segmentation, AudioCLIP for vision-audio feature alignment, and a pre-trained Stable Audio model for high-fidelity sound generation (see Fig. \ref{fig_framework}).

\textbf{Object Segmentation with SAM.}
Some objects produce the same sound when hit at different locations,
while many objects consist of multiple components that produce distinct sounds when struck.
Thus, a key challenge in sound prediction is ensuring that the model focuses on the correct region.
To address this, we leverage the Segment Anything Model (SAM)~\cite{sam} to segment images at multiple levels, including subpart-level, part-level, and whole-object-level.
Then, the hitting region can be localized by selecting the object segmentation (at multi-level) where the hitting position lies.
This hierarchical segmentation allows our model to better capture and localize the hitting regions for various objects.

\textbf{AudioCLIP for Vision-Audio Feature Alignment.}
Even with precise segmentation, predicting plausible sounds requires a meaningful feature representation that bridges the visual and auditory domains. Traditional feature extractors, such as CLIP~\cite{clip}, are optimized for image-text alignment but lack audio-specific embeddings. To overcome this, we employ AudioCLIP~\cite{audioclip}, which extends CLIP by incorporating audio representations alongside vision and text. 
AudioCLIP enables the extraction of semantically rich, multimodal embeddings that align visual textures with their corresponding sound characteristics. Moreover, AudioCLIP supports zero-shot generalization, allowing the model to fit diverse real-world scenes.
In practice, we use AudioCLIP to extract the features of the multi-level object segmentations where hitting is performed.
Building on this, we follow \cite{langsplat} to construct a feature-augmented 3DGS, enabling 3D view-consistent visual features at any location in the scene.

\textbf{Pre-Trained Stable Audio for Sound Generation.}
We then use a model $M$ to predict the sound $x$ at a hitting location based on the corresponding multi-level features $\{f^s, f^p, f^w\}$ as $M(x| \{f^s, f^p, f^w\})$.
However, generating high-quality, realistic impact sounds from a limited dataset is a major challenge, as training such a model from scratch would require an extensive dataset of diverse impact sounds. To address this, we fine-tune a pre-trained Stable Audio model~\cite{stableaudioopen}, a state-of-the-art and text-conditioned audio generation model based on the diffusion transformer.
This pre-trained Stable Audio offers several advantages, including enabling generalization to diverse objects by retaining prior knowledge about a wide range of impact sounds and achieving training efficiency by fine-tuning only a small part of the model.
Specifically, we replace its original conditioning mechanism with a multi-level visual feature conditioner and shorten the generated sample length to match the duration of a hitting sound. Only the visual feature conditioner is trained during fine-tuning, while the transformer weights remain frozen to preserve the model’s generalization ability.

\textbf{Inference Pipeline.}
During inference, for any queried impact location, we extract multi-scale SAM segmentation features and AudioCLIP embeddings from the feature 3DGS rendering. These embeddings are then passed as conditions to the Stable Audio model, which synthesizes the corresponding impact sound.

\subsection{Localizing Sound in 3D}
To better understand the relationship between vision and sound, we also introduce the task of sound localization: given a hitting sound, the goal is to predict the location in the scene that produced it (see Fig. \ref{fig_framework}). To achieve this, we first train AudioCLIP \cite{audioclip}, a cross-modal visual-sound encoder, using self-supervised contrastive learning on our training dataset, drawing inspiration from the text-image contrastive pre-training commonly used in image generation \cite{clip}. We then leverage this visual-audio encoder and SAM \cite{sam} to encode multiview images of each scene and extract visual features. These features are subsequently used to train the feature-augmented 3DGS \cite{langsplat} for each scene. After we have the feature-augmented 3DGS, we can query it with sound to localize the sound's origin.

\textbf{Inference Pipeline.} Given a hitting sound and a viewpoint, we first render the feature-augmented 3DGS from the specified viewpoint to obtain the rendered visual embedding $\phi_{i} \in \mathbb{R}^{(H \times W) \times d}$, where $d$ is the embedding dimension. We then encode the hitting sound using the fine-tuned AudioCLIP to extract the audio features $\phi_{s} \in \mathbb{R}^{1 \times d}$. Next, we compute the relevance score using a dot product operation, $\phi_{i} \cdot \phi_{s} \in \mathbb{R}^{(H \times W) \times 1}$. The region with a higher relevance score will be predicted with greater confidence as the localization result.
\section{Dataset Statistics}
\begin{figure*}[t]
    \includegraphics[width=1.0\linewidth]{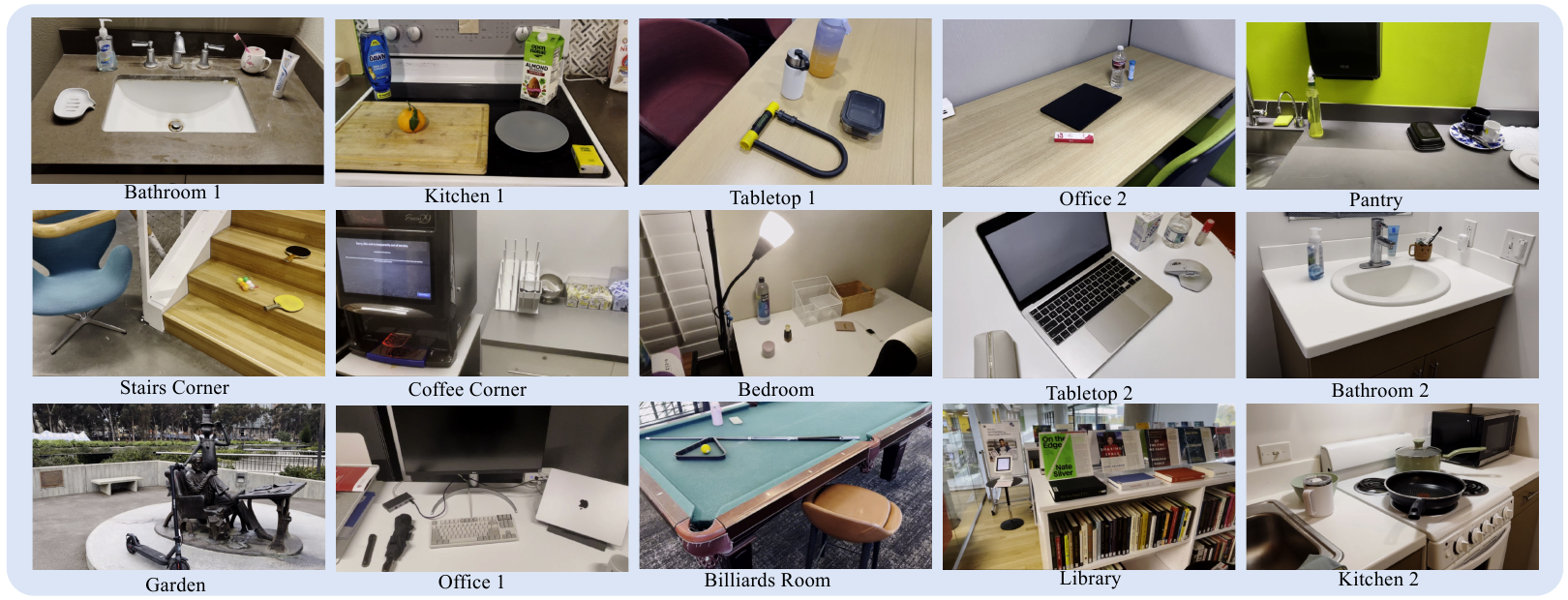}
    \caption{
    \textbf{Example scenes in our dataset.} Our dataset consists of 15 diverse environments, including indoor and outdoor settings such as a bedroom, kitchen, bathroom, office, library, coffee corner, and garden. Each scene contains various materials (e.g., wood, metal, plastic, ceramic) and impact locations, yielding a rich collection of visual-audio pairs for training and evaluation.
    }
    \label{fig_scene_samples}
\end{figure*}

We collected data from 15 different scenes, including a bedroom, kitchen, bathroom, office, library, tabletop, various corners in a teaching building, etc. The sound sources in our dataset include materials such as wood, ceramic, plastic, metal, LCDs, etc. Each scene contains between 100 and 200 data points, depending on the diversity of sound sources present. In total, our dataset comprises approximately 2,000 visual-sound data pairs. Fig. \ref{fig_scene_samples} provides examples of some of the collected scenes, and we include all 15 scene images in our supplementary materials.

\begin{table}[ht]
    \centering
    \renewcommand\tabcolsep{3.75pt}
    \begin{tabular}{l|c c c}
        \toprule
         Dataset & Scenario & Source & 3D\\
         \midrule
         ObjectFolder \cite{gao2021objectfolder} & Object & Synthetic & \checkmark\\
         \midrule
         ObjectFolder 2.0 \cite{gao2022objectfolder} & Object & Synthetic & \checkmark\\
         \midrule
         ObjectFolder Real \cite{gao2023objectfolder} & Object & Robot & \checkmark\\
         \midrule
         Visually Indicated Sound \cite{owens2016visually} & Scene & Human & $\times$\\
         \midrule
         AudioSet \cite{audioset} & Scene & Internet & $\times$ \\
         \midrule
         \textcolor{teal}{Ours} & Scene & Human & \checkmark\\
         \bottomrule
    \end{tabular}
    \caption{
    We compare our dataset with existing hitting sound datasets in terms of scenario (object-level vs. scene-level), data source (synthesized by simulators, collected by robot, or human-collected), and whether the dataset contains 3D spatial information. Unlike prior datasets focusing on object-level interactions or synthetic environments, our dataset captures real-world, human-collected impact sounds at the scene level with full 3D spatial alignment.
    }
    \label{tab:data_comparison}
\end{table}

\section{Experiments}
We conduct all experiments on our collected datasets. In all experimental settings, we adopt a 4:1 train-test split ratio.

\subsection{Hitting Sound Generation.}
To evaluate the quality of our hitting sound generation, we compare our model with im2wav \cite{sheffer2023iheartruecolors}, the only open-sourced image-to-audio generation model available retrained on our dataset. Additionally, we investigate the impact of using object-level features versus local texture features around the hitting positions as conditioning inputs for sound inference. 

In Tab. \ref{tab:sound_gen_result}, im2wav \cite{sheffer2023iheartruecolors} refers to the model conditioned on object segments obtained from SAM \cite{sam}, while im2wav (local) uses 100 $\times$ 100 rendered images centered at the hitting locations as the input. Ours represents our approach, which conditions the model on multi-level object features extracted from SAM \cite{sam} and CLIP \cite{audioclip}. In contrast, Ours (local) conditions the model on local features extracted from 100 $\times$ 100 rendered images centered at the hitting position using ResNet-34 \cite{resnet}. The metric we use includes Frechet Audio Distance (FAD) \cite{fad}, Kullback-Leibler Divergence (KL), Structural Similarity Index Measure (SSIM), and Peak Signal-to-Noise Ratio (PSNR). Similarly to the Frechet Distance in image generation, FAD is widely used in audio generation to measure the distance between the generated and real distributions. KL is computed at the paired sample level, then summed and averaged to obtain the final result. Our method achieves lower FAD and KL scores compared to others, indicating that our proposed model more effectively captures the distributions and characteristics of the hitting sound data. At a low level, our method also attains higher SSIM and PSNR scores, demonstrating that our generated results more faithfully resemble their ground truth counterparts.

Besides computing sound metrics, we also evaluate the quality of our generated sounds through a survey, as we believe human perception is always the \textit{gold standard} for evaluation. In our survey, we randomly select 50 hitting location images from our test set, ensuring a diverse range of hitting materials. For each location, we provide the ground truth hitting sound we collected, the sound generated by our method, and the sound generated by im2wav \cite{sheffer2023iheartruecolors}. Participants are then asked to choose the sound they believe best matches the hitting sound at the given location. We collected 30 responses through a Google Form, and the results are shown in Tab. \ref{tab:sound_gen_survey}. Among the total $30 \times 50 = 1500$ survey data points, 42.93\% of real sounds and 41.93\% of our generated sounds were selected as the ones that best matched the hitting location, which indicates that our generated sounds are almost indistinguishable from real sounds.

\begin{table}[t]
    \centering
    \begin{tabular}{l|c c c c}
        \toprule
         Method & FAD $\downarrow$ & SSIM $\uparrow$ & PSNR $\uparrow$ & KL $\downarrow$ \\
         \midrule
         im2wav (local) & 1.50 & 0.67 & 14.65 & 0.52\\
         \midrule
         im2wav \cite{sheffer2023iheartruecolors} & 1.68 & 0.67 & 15.03 & 0.48\\
        \midrule
        Ours (local) & 0.66 & 0.77 & 18.68 & 0.43\\
        \midrule
         Ours & \textcolor{teal}{0.35} & \textcolor{teal}{0.82} & \textcolor{teal}{20.83} & \textcolor{teal}{0.38}\\
         \bottomrule
    \end{tabular}
    \caption{
    \textbf{Sound Generation Results.} We compare our method with im2wav~\cite{sheffer2023iheartruecolors} and several variants with various evaluation metrics. 
    Our approach achieves the best results across all metrics, demonstrating its ability to generate impact sounds that closely match real-world recordings.
    }
    \label{tab:sound_gen_result}
\end{table}

\begin{table}[ht]
    \centering
    \renewcommand\tabcolsep{3.75pt}
    \begin{tabular}{l|c c c}
        \toprule
         & Ground-truth & Ours & Im2wav \cite{sheffer2023iheartruecolors}\\
         \midrule
         Best Match & 42.93 \% & 41.93 \% & 15.13\%\\
         \bottomrule
    \end{tabular}
    \caption{
    \textbf{Sound Generation User Study.} We conducted a user study to evaluate the perceptual quality of generated impact sounds. 
    Participants were presented with images of impact locations along with three sound samples: the ground-truth recorded sound, our generated sound, and the sound generated by im2wav~\cite{sheffer2023iheartruecolors}. They were asked to select the sound that best matched the given impact location. 
    The results show that our method produces nearly indistinguishable sounds from real impact sounds.
    }
    \label{tab:sound_gen_survey}
\end{table}

\begin{figure}[ht]
    \includegraphics[width=1.0\linewidth]{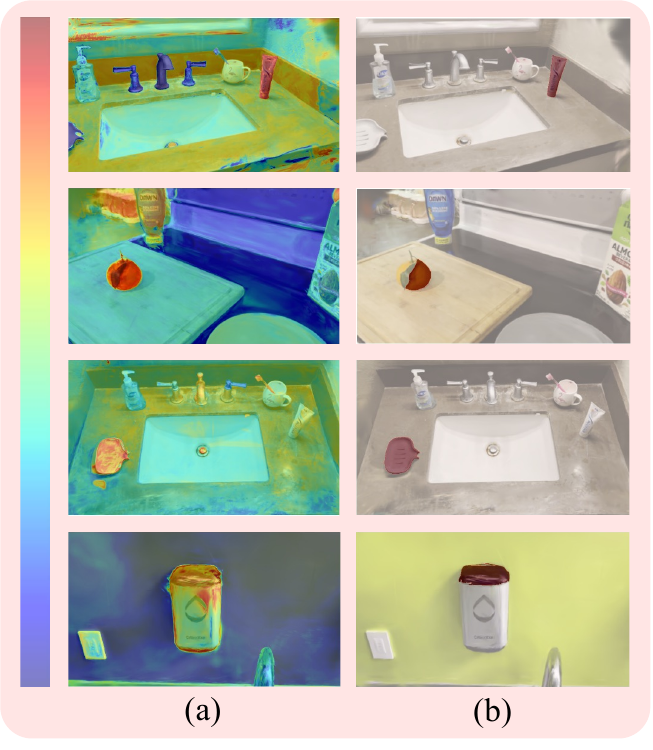}
    \caption{
    Visualization of sound localization results. Given an input hitting sound, our model predicts the most relevant impact location within the 3D scene.
    (a) The heatmap represents the localization confidence scores, where brighter regions indicate higher confidence for the predicted sound source. 
    (b) The highlighted region denotes the final localized impact objects (or parts).
    }
\label{fig_localization}
\end{figure}

\subsection{Hitting Sound Localization.}

To quantitatively evaluate our method, we first use SAM \cite{sam} (default scale) to segment the rendered scene from 3DGS. Next, we encode each segmented image along with the given hitting audio using our fine-tuned AudioCLIP \cite{audioclip}. Finally, we compute the relevance score between the query audio and each segmented part, and the part with the highest relevance score is chosen as the localization result.

We compare our method with AudioCLIP without fine-tuning, which utilizes the AudioCLIP model trained on the large-scale AudioSet \cite{audioset}, and with Random, which selects a segment in the scene at random as the localization result. We manually label each result as correct or incorrect, as a single sound can sometimes correspond to multiple segments identified by SAM. For example, in the bathroom scene shown in Fig. \ref{fig_scene_samples}, the metal faucet is segmented into three parts (left, middle, and right) by SAM. However, since all three segments produce the same sound, we consider the result correct if a faucet sound is mapped to any of these segments.

As shown in Table \ref{tab:sound_loc_result}, we measure Acc(1) and Acc(3). Acc(1) represents the accuracy rate when the segment with the highest relevance score is the correct result, while Acc(3) indicates the accuracy rate when the correct location is covered by one of the top three segments with the highest relevance scores. Our method significantly outperforms the baselines in both top-1 and top-3 localization accuracy rates. 
This suggests that existing large-scale sound datasets, such as AudioSet \cite{audioset}, lack a sufficient number of clean and high quality hitting sound samples to train a contrastive learning model and effectively perform localization tasks, despite the prevalence of hitting sounds in everyday life. With our collected dataset and proposed pipeline, we can effectively localize sound in a 3D scene.

In Fig. \ref{fig_localization}, we present a visualization of the sound localization results. The left column shows the heatmap generated when using the hitting sound to query rendered features at every pixel. The right column highlights the segmented region selected by our method. These results demonstrate that our approach effectively identifies impact locations with high accuracy, even in complex environments.

\begin{figure}
    \centering
    \includegraphics[width=1\linewidth]{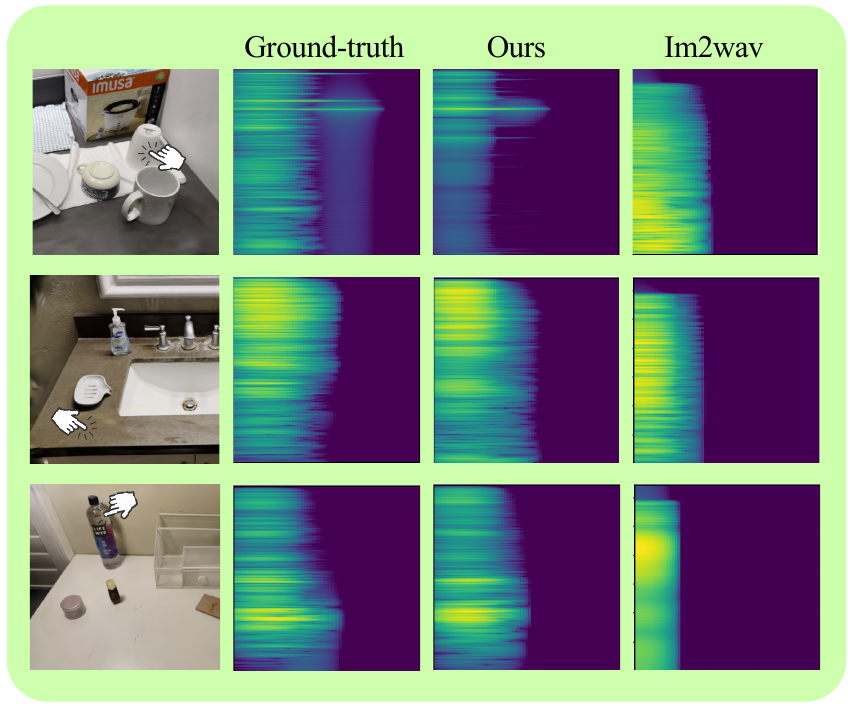}
    \caption{Mel Spectrogram of generated sounds. We compare the audio generated by our method with ground-truth recordings and results from im2wav~\cite{sheffer2023hear}. The visualized spectrograms of sounds show that our approach produces impact sounds that closely resemble real recordings.}
    \label{fig_sound_vis}
\end{figure}

In Fig. \ref{fig_sound_vis}, we present visualizations of our generated sound, the baseline, and the groundtruth sound. The Mel spectrogram reveals that our generated sounds exhibit similar frequency content and temporal evolution compared with the groundtruth sound. This aligns with our survey results, which indicate that our generated sound is nearly indistinguishable from the ground truth sound.

\begin{table}[t]
    \centering 
    \renewcommand\tabcolsep{3.75pt}
    \begin{tabular}{l|c c c}
        \toprule
           & AudioClIP \cite{audioclip} &   FT-AudioClIP (ours)& Random\\
         \midrule
         Acc(1) &  19.0\% & \textcolor{teal}{74.4\%} & 10.2\%\\
         Acc(3) & 35.9\% & \textcolor{teal}{85.5\%} & 26.8\%\\
         \bottomrule
    \end{tabular}
    \caption{
    \textbf{Sound Localization Accuracy.} We compare our method (FT-AudioCLIP) with the baseline AudioCLIP~\cite{audioclip} 
    (without fine-tuning) and the random selection. \textbf{Acc(1)} denotes the accuracy when the top-1 predicted location is correct, while \textbf{Acc(3)} 
    represents the accuracy when the correct location is within the top-3 predictions. Our fine-tuned model significantly outperforms both baselines.
    }
    \label{tab:sound_loc_result}
\end{table}
\subsection{Implementation Details}

\textbf{RGB and Feature Field.} We follow the official 3DGS implementation \cite{3dgs} to render RGB images. For feature extraction, we use the Langsplat codebase \cite{langsplat} to obtain multilevel SAM features and reconstruct the feature fields. During data collection, we capture a video for each scene and uniformly sample approximately 300 frames from it as the training set for scene reconstruction. We train both the RGB fields and feature fields on a single NVIDIA GeForce RTX 3090 GPU.

\textbf{Sound Generation and Localization.} For sound generation, our implementation builds upon Stable Audio Open \cite{stableaudioopen}, using its released checkpoint as the starting point for training. We modify its conditioning encoder to process visual features by incorporating MLPs. During training, we optimize the conditioning encoder using the AdamW optimizer with a base learning rate of $5e^{-5}$ and a batch size of 8 on a single NVIDIA A100 GPU. For inference, we apply classifier-free guidance with a scale of 6 and perform 250 sampling steps. To evaluate sound quality, we use the codebase from \cite{audioldm2}. For sound localization, we train AudioClIP \cite{audioclip} on our dataset using an SGD optimizer with a learning rate of $5e^{-5}$ on a single NVIDIA GeForce RTX 3090 GPU. We initialize the image encoder with pretrained CLIP \cite{clip} weights and freeze it during training. Only the weights of audio encoder part are updated.

\textbf{Baseline.} For the baseline, we use the open-sourced implementation of im2wav~\cite{sheffer2023iheartruecolors}. Since, by default, the model cannot handle very short sequences, such as our hitting sounds, we pad them to 10 seconds for training and inference, then crop them afterward for metric calculation. When training im2wav, we use a batch size of 16 for the VQVAE and a batch size of 8 for the upsampling and low-level models. All baseline training is done on a single NVIDIA GeForce RTX 3090 GPU.
\section{Conclusion, Limitation, and Future Work}
\label{sec:conclusion}

In this paper, we introduce Visual Acoustic Fields, a novel framework that integrates visual and acoustic signals within a 3D scene. Our approach leverages 3D Gaussian Splatting to establish a spatially consistent 3D scene representation for interactive sound generation.
By incorporating Audio-CLIP features, our model enables both vision-conditioned sound generation and sound localization. To support this research, we propose a re-rendering strategy for dataset collection, providing sound annotations with accurate impact locations. Extensive experiments on our newly curated dataset demonstrate the effectiveness of our approach in generating plausible hitting sounds and accurately localizing sound sources in 3D space.

\textbf{Limitations and Future Works.}
Despite the promising results, our approach has several limitations. First, our dataset, while diverse, remains limited to static scenes and a variety of materials. Expanding the dataset to include more environments, such as including dynamic scenes and more outdoor settings, and objects with richer acoustic properties would enable better generalization of our model.
Another limitation of our Visual Acoustic Fields model is its agnosticism of the spatial location of the listener. 
Specifically, our model does not account for the perceived sound changes based on the listener’s position.
To overcome this limitation, future research could incorporate physics-based sound propagation models that account for distance attenuation, occlusion, and echo.

{
    \small
    \bibliographystyle{ieeenat_fullname}
    \bibliography{main}

\begin{thebibliography}{50}
\providecommand{\natexlab}[1]{#1}
\providecommand{\url}[1]{\texttt{#1}}
\expandafter\ifx\csname urlstyle\endcsname\relax
  \providecommand{\doi}[1]{doi: #1}\else
  \providecommand{\doi}{doi: \begingroup \urlstyle{rm}\Url}\fi

\bibitem[Afouras et~al.(2018)Afouras, Chung, Senior, Vinyals, and Zisserman]{afouras2018deep}
Triantafyllos Afouras, Joon~Son Chung, Andrew Senior, Oriol Vinyals, and Andrew Zisserman.
\newblock Deep audio-visual speech recognition.
\newblock \emph{IEEE transactions on pattern analysis and machine intelligence}, 44\penalty0 (12):\penalty0 8717--8727, 2018.

\bibitem[Arandjelovic and Zisserman(2018)]{arandjelovic2018objects}
Relja Arandjelovic and Andrew Zisserman.
\newblock Objects that sound.
\newblock In \emph{Proceedings of the European conference on computer vision (ECCV)}, pages 435--451, 2018.

\bibitem[Baillargeon(2002)]{baillargeon2002acquisition}
Ren{\'e}e Baillargeon.
\newblock The acquisition of physical knowledge in infancy: A summary in eight lessons.
\newblock \emph{Blackwell handbook of childhood cognitive development}, pages 47--83, 2002.

\bibitem[Calandra et~al.(2017)Calandra, Owens, Upadhyaya, Yuan, Lin, Adelson, and Levine]{calandra2017feeling}
Roberto Calandra, Andrew Owens, Manu Upadhyaya, Wenzhen Yuan, Justin Lin, Edward~H Adelson, and Sergey Levine.
\newblock The feeling of success: Does touch sensing help predict grasp outcomes?
\newblock \emph{arXiv preprint arXiv:1710.05512}, 2017.

\bibitem[Calandra et~al.(2018)Calandra, Owens, Jayaraman, Lin, Yuan, Malik, Adelson, and Levine]{calandra2018more}
Roberto Calandra, Andrew Owens, Dinesh Jayaraman, Justin Lin, Wenzhen Yuan, Jitendra Malik, Edward~H Adelson, and Sergey Levine.
\newblock More than a feeling: Learning to grasp and regrasp using vision and touch.
\newblock \emph{IEEE Robotics and Automation Letters}, 3\penalty0 (4):\penalty0 3300--3307, 2018.

\bibitem[Chen et~al.(2024)Chen, Peng, Baid, Xue, Hsu, Harwath, and Grauman]{chen2024action2sound}
Changan Chen, Puyuan Peng, Ami Baid, Zihui Xue, Wei-Ning Hsu, David Harwath, and Kristen Grauman.
\newblock Action2sound: Ambient-aware generation of action sounds from egocentric videos.
\newblock In \emph{European Conference on Computer Vision}, pages 277--295. Springer, 2024.

\bibitem[Chung and Zisserman(2017)]{chung2017lip}
Joon~Son Chung and Andrew Zisserman.
\newblock Lip reading in the wild.
\newblock In \emph{Computer Vision--ACCV 2016: 13th Asian Conference on Computer Vision, Taipei, Taiwan, November 20-24, 2016, Revised Selected Papers, Part II 13}, pages 87--103. Springer, 2017.

\bibitem[Dou et~al.(2024)Dou, Yang, Liu, Loquercio, and Owens]{tactileaugmentedradiancefields}
Yiming Dou, Fengyu Yang, Yi Liu, Antonio Loquercio, and Andrew Owens.
\newblock Tactile-augmented radiance fields, 2024.

\bibitem[Evans et~al.(2024)Evans, Parker, Carr, Zukowski, Taylor, and Pons]{stableaudioopen}
Zach Evans, Julian~D. Parker, CJ Carr, Zack Zukowski, Josiah Taylor, and Jordi Pons.
\newblock Stable audio open, 2024.

\bibitem[Gan et~al.(2020)Gan, Huang, Chen, Tenenbaum, and Torralba]{gan2020foley}
Chuang Gan, Deng Huang, Peihao Chen, Joshua~B Tenenbaum, and Antonio Torralba.
\newblock Foley music: Learning to generate music from videos.
\newblock In \emph{Computer Vision--ECCV 2020: 16th European Conference, Glasgow, UK, August 23--28, 2020, Proceedings, Part XI 16}, pages 758--775. Springer, 2020.

\bibitem[Gao et~al.(2021)Gao, Chang, Mall, Fei-Fei, and Wu]{gao2021objectfolder}
Ruohan Gao, Yen-Yu Chang, Shivani Mall, Li Fei-Fei, and Jiajun Wu.
\newblock Objectfolder: A dataset of objects with implicit visual, auditory, and tactile representations.
\newblock In \emph{Conference on Robot Learning}, 2021.

\bibitem[Gao et~al.(2022)Gao, Si, Chang, Clarke, Bohg, Fei-Fei, Yuan, and Wu]{gao2022objectfolder}
Ruohan Gao, Zilin Si, Yen-Yu Chang, Samuel Clarke, Jeannette Bohg, Li Fei-Fei, Wenzhen Yuan, and Jiajun Wu.
\newblock Objectfolder 2.0: A multisensory object dataset for sim2real transfer.
\newblock In \emph{Proceedings of the IEEE/CVF conference on computer vision and pattern recognition}, pages 10598--10608, 2022.

\bibitem[Gao et~al.(2023)Gao, Dou, Li, Agarwal, Bohg, Li, Fei-Fei, and Wu]{gao2023objectfolder}
Ruohan Gao, Yiming Dou, Hao Li, Tanmay Agarwal, Jeannette Bohg, Yunzhu Li, Li Fei-Fei, and Jiajun Wu.
\newblock The objectfolder benchmark: Multisensory learning with neural and real objects.
\newblock In \emph{Proceedings of the IEEE/CVF Conference on Computer Vision and Pattern Recognition}, pages 17276--17286, 2023.

\bibitem[Gemmeke et~al.(2017)Gemmeke, Ellis, Freedman, Jansen, Lawrence, Moore, Plakal, and Ritter]{audioset}
Jort~F. Gemmeke, Daniel P.~W. Ellis, Dylan Freedman, Aren Jansen, Wade Lawrence, R.~Channing Moore, Manoj Plakal, and Marvin Ritter.
\newblock Audio set: An ontology and human-labeled dataset for audio events.
\newblock In \emph{2017 IEEE International Conference on Acoustics, Speech and Signal Processing (ICASSP)}, pages 776--780, 2017.

\bibitem[Guzhov et~al.(2021)Guzhov, Raue, Hees, and Dengel]{audioclip}
Andrey Guzhov, Federico Raue, Jörn Hees, and Andreas Dengel.
\newblock Audioclip: Extending clip to image, text and audio, 2021.

\bibitem[He et~al.(2015)He, Zhang, Ren, and Sun]{resnet}
Kaiming He, Xiangyu Zhang, Shaoqing Ren, and Jian Sun.
\newblock Deep residual learning for image recognition, 2015.

\bibitem[Hong et~al.(2024)Hong, Zheng, Chen, Wang, Li, and Gan]{hong2024multiply}
Yining Hong, Zishuo Zheng, Peihao Chen, Yian Wang, Junyan Li, and Chuang Gan.
\newblock Multiply: A multisensory object-centric embodied large language model in 3d world.
\newblock In \emph{Proceedings of the IEEE/CVF Conference on Computer Vision and Pattern Recognition}, pages 26406--26416, 2024.

\bibitem[Huang et~al.(2023{\natexlab{a}})Huang, Mees, Zeng, and Burgard]{huang2023audio}
Chenguang Huang, Oier Mees, Andy Zeng, and Wolfram Burgard.
\newblock Audio visual language maps for robot navigation.
\newblock In \emph{International Symposium on Experimental Robotics}, pages 105--117. Springer, 2023{\natexlab{a}}.

\bibitem[Huang et~al.(2023{\natexlab{b}})Huang, Li, Wang, Zhu, Dai, Han, Rong, and Liu]{huang2023discovering}
Shaofei Huang, Han Li, Yuqing Wang, Hongji Zhu, Jiao Dai, Jizhong Han, Wenge Rong, and Si Liu.
\newblock Discovering sounding objects by audio queries for audio visual segmentation.
\newblock \emph{arXiv preprint arXiv:2309.09501}, 2023{\natexlab{b}}.

\bibitem[Isola et~al.(2015)Isola, Zoran, Krishnan, and Adelson]{isola2015learningvisualgroupscooccurrences}
Phillip Isola, Daniel Zoran, Dilip Krishnan, and Edward~H. Adelson.
\newblock Learning visual groups from co-occurrences in space and time, 2015.

\bibitem[Jatavallabhula et~al.(2023)Jatavallabhula, Kuwajerwala, Gu, Omama, Chen, Li, Iyer, Saryazdi, Keetha, Tewari, Tenenbaum, {de Melo}, Krishna, Paull, Shkurti, and Torralba]{conceptfusion}
{Krishna Murthy} Jatavallabhula, Alihusein Kuwajerwala, Qiao Gu, Mohd Omama, Tao Chen, Shuang Li, Ganesh Iyer, Soroush Saryazdi, Nikhil Keetha, Ayush Tewari, {Joshua B.} Tenenbaum, {Celso Miguel} {de Melo}, Madhava Krishna, Liam Paull, Florian Shkurti, and Antonio Torralba.
\newblock Conceptfusion: Open-set multimodal 3d mapping.
\newblock \emph{Robotics: Science and Systems (RSS)}, 2023.

\bibitem[Kerbl et~al.(2023{\natexlab{a}})Kerbl, Kopanas, Leimk{\"u}hler, and Drettakis]{kerbl20233d}
Bernhard Kerbl, Georgios Kopanas, Thomas Leimk{\"u}hler, and George Drettakis.
\newblock 3d gaussian splatting for real-time radiance field rendering.
\newblock \emph{ACM Trans. Graph.}, 42\penalty0 (4):\penalty0 139--1, 2023{\natexlab{a}}.

\bibitem[Kerbl et~al.(2023{\natexlab{b}})Kerbl, Kopanas, Leimkühler, and Drettakis]{3dgs}
Bernhard Kerbl, Georgios Kopanas, Thomas Leimkühler, and George Drettakis.
\newblock 3d gaussian splatting for real-time radiance field rendering, 2023{\natexlab{b}}.

\bibitem[Kerr et~al.(2023)Kerr, Huang, Wilcox, Hoque, Ichnowski, Calandra, and Goldberg]{kerr2022self}
Justin Kerr, Huang Huang, Albert Wilcox, Ryan Hoque, Jeffrey Ichnowski, Roberto Calandra, and Ken Goldberg.
\newblock Self-supervised visuo-tactile pretraining to locate and follow garment features.
\newblock \emph{Robotics: Science and Systems (RSS)}, 2023.

\bibitem[Kilgour et~al.(2019)Kilgour, Zuluaga, Roblek, and Sharifi]{fad}
Kevin Kilgour, Mauricio Zuluaga, Dominik Roblek, and Matthew Sharifi.
\newblock Fr\'echet audio distance: A metric for evaluating music enhancement algorithms, 2019.

\bibitem[Kirillov et~al.(2023)Kirillov, Mintun, Ravi, Mao, Rolland, Gustafson, Xiao, Whitehead, Berg, Lo, Dollár, and Girshick]{sam}
Alexander Kirillov, Eric Mintun, Nikhila Ravi, Hanzi Mao, Chloe Rolland, Laura Gustafson, Tete Xiao, Spencer Whitehead, Alexander~C. Berg, Wan-Yen Lo, Piotr Dollár, and Ross Girshick.
\newblock Segment anything, 2023.

\bibitem[Li et~al.(2021)Li, Yang, Ross, and Kanazawa]{li2021ai}
Ruilong Li, Shan Yang, David~A Ross, and Angjoo Kanazawa.
\newblock Ai choreographer: Music conditioned 3d dance generation with aist++.
\newblock In \emph{Proceedings of the IEEE/CVF International Conference on Computer Vision}, pages 13401--13412, 2021.

\bibitem[Li et~al.(2019)Li, Zhu, Tedrake, and Torralba]{li2019connecting}
Yunzhu Li, Jun-Yan Zhu, Russ Tedrake, and Antonio Torralba.
\newblock Connecting touch and vision via cross-modal prediction.
\newblock In \emph{Proceedings of the IEEE/CVF Conference on Computer Vision and Pattern Recognition}, pages 10609--10618, 2019.

\bibitem[Liu et~al.(2024)Liu, Yuan, Liu, Mei, Kong, Tian, Wang, Wang, Wang, and Plumbley]{audioldm2}
Haohe Liu, Yi Yuan, Xubo Liu, Xinhao Mei, Qiuqiang Kong, Qiao Tian, Yuping Wang, Wenwu Wang, Yuxuan Wang, and Mark~D. Plumbley.
\newblock Audioldm 2: Learning holistic audio generation with self-supervised pretraining.
\newblock \emph{IEEE/ACM Transactions on Audio, Speech, and Language Processing}, 32:\penalty0 2871--2883, 2024.

\bibitem[Luo et~al.(2023)Luo, Yan, Hu, and Zhao]{luo2023difffoleysynchronizedvideotoaudiosynthesis}
Simian Luo, Chuanhao Yan, Chenxu Hu, and Hang Zhao.
\newblock Diff-foley: Synchronized video-to-audio synthesis with latent diffusion models, 2023.

\bibitem[Minderer et~al.(2023)Minderer, Gritsenko, and Houlsby]{minderer2023scaling}
Matthias Minderer, Alexey Gritsenko, and Neil Houlsby.
\newblock Scaling open-vocabulary object detection.
\newblock \emph{Advances in Neural Information Processing Systems}, 36:\penalty0 72983--73007, 2023.

\bibitem[Owens et~al.(2016)Owens, Isola, McDermott, Torralba, Adelson, and Freeman]{owens2016visually}
Andrew Owens, Phillip Isola, Josh McDermott, Antonio Torralba, Edward~H Adelson, and William~T Freeman.
\newblock Visually indicated sounds.
\newblock In \emph{Proceedings of the IEEE conference on computer vision and pattern recognition}, pages 2405--2413, 2016.

\bibitem[Piczak(2015)]{piczak2015esc}
Karol~J Piczak.
\newblock Esc: Dataset for environmental sound classification.
\newblock In \emph{Proceedings of the 23rd ACM international conference on Multimedia}, pages 1015--1018, 2015.

\bibitem[Qin et~al.(2024)Qin, Li, Zhou, Wang, and Pfister]{langsplat}
Minghan Qin, Wanhua Li, Jiawei Zhou, Haoqian Wang, and Hanspeter Pfister.
\newblock Langsplat: 3d language gaussian splatting, 2024.

\bibitem[Radford et~al.(2021)Radford, Kim, Hallacy, Ramesh, Goh, Agarwal, Sastry, Askell, Mishkin, Clark, Krueger, and Sutskever]{clip}
Alec Radford, Jong~Wook Kim, Chris Hallacy, Aditya Ramesh, Gabriel Goh, Sandhini Agarwal, Girish Sastry, Amanda Askell, Pamela Mishkin, Jack Clark, Gretchen Krueger, and Ilya Sutskever.
\newblock Learning transferable visual models from natural language supervision, 2021.

\bibitem[Salamon et~al.(2014)Salamon, Jacoby, and Bello]{UrbanSound}
J. Salamon, C. Jacoby, and J.~P. Bello.
\newblock A dataset and taxonomy for urban sound research.
\newblock In \emph{22nd {ACM} International Conference on Multimedia (ACM-MM'14)}, pages 1041--1044, Orlando, FL, USA, 2014.

\bibitem[Sch\"{o}nberger and Frahm(2016)]{schoenberger2016sfm}
Johannes~Lutz Sch\"{o}nberger and Jan-Michael Frahm.
\newblock Structure-from-motion revisited.
\newblock In \emph{Conference on Computer Vision and Pattern Recognition (CVPR)}, 2016.

\bibitem[Sch\"{o}nberger et~al.(2016)Sch\"{o}nberger, Zheng, Pollefeys, and Frahm]{schoenberger2016mvs}
Johannes~Lutz Sch\"{o}nberger, Enliang Zheng, Marc Pollefeys, and Jan-Michael Frahm.
\newblock Pixelwise view selection for unstructured multi-view stereo.
\newblock In \emph{European Conference on Computer Vision (ECCV)}, 2016.

\bibitem[Schulz(2012)]{originofinquiry}
Laura Schulz.
\newblock The origins of inquiry: inductive inference and exploration in early childhood.
\newblock \emph{Trends in Cognitive Sciences}, 16\penalty0 (7):\penalty0 382--389, 2012.

\bibitem[Sheffer and Adi(2023{\natexlab{a}})]{sheffer2023hear}
Roy Sheffer and Yossi Adi.
\newblock I hear your true colors: Image guided audio generation.
\newblock In \emph{ICASSP 2023-2023 IEEE International Conference on Acoustics, Speech and Signal Processing (ICASSP)}, pages 1--5. IEEE, 2023{\natexlab{a}}.

\bibitem[Sheffer and Adi(2023{\natexlab{b}})]{sheffer2023iheartruecolors}
Roy Sheffer and Yossi Adi.
\newblock I hear your true colors: Image guided audio generation, 2023{\natexlab{b}}.

\bibitem[Su et~al.(2023)Su, Qian, Shlizerman, Torralba, and Gan]{su2023physics}
Kun Su, Kaizhi Qian, Eli Shlizerman, Antonio Torralba, and Chuang Gan.
\newblock Physics-driven diffusion models for impact sound synthesis from videos.
\newblock In \emph{Proceedings of the IEEE/CVF Conference on Computer Vision and Pattern Recognition}, pages 9749--9759, 2023.

\bibitem[Wang et~al.(2023)Wang, Ma, Pascual, Cartwright, and Cai]{v2amapper}
Heng Wang, Jianbo Ma, Santiago Pascual, Richard Cartwright, and Weidong Cai.
\newblock V2a-mapper: A lightweight solution for vision-to-audio generation by connecting foundation models, 2023.

\bibitem[Wang et~al.(2020)Wang, Wu, Song, Yang, Wu, Qian, He, Qiao, and Loy]{wang2020mead}
Kaisiyuan Wang, Qianyi Wu, Linsen Song, Zhuoqian Yang, Wayne Wu, Chen Qian, Ran He, Yu Qiao, and Chen~Change Loy.
\newblock Mead: A large-scale audio-visual dataset for emotional talking-face generation.
\newblock In \emph{European Conference on Computer Vision}, pages 700--717. Springer, 2020.

\bibitem[Xing et~al.(2024)Xing, He, Tian, Wang, and Chen]{xing2024seeing}
Yazhou Xing, Yingqing He, Zeyue Tian, Xintao Wang, and Qifeng Chen.
\newblock Seeing and hearing: Open-domain visual-audio generation with diffusion latent aligners.
\newblock In \emph{Proceedings of the IEEE/CVF Conference on Computer Vision and Pattern Recognition}, pages 7151--7161, 2024.

\bibitem[Zhang et~al.(2024)Zhang, Gu, Zeng, Xing, Wang, Wu, and Chen]{zhang2024foleycrafter}
Yiming Zhang, Yicheng Gu, Yanhong Zeng, Zhening Xing, Yuancheng Wang, Zhizheng Wu, and Kai Chen.
\newblock Foleycrafter: Bring silent videos to life with lifelike and synchronized sounds.
\newblock \emph{arXiv preprint arXiv:2407.01494}, 2024.

\bibitem[Zhao et~al.(2019)Zhao, Gan, Ma, and Torralba]{zhao2019sound}
Hang Zhao, Chuang Gan, Wei-Chiu Ma, and Antonio Torralba.
\newblock The sound of motions.
\newblock In \emph{Proceedings of the IEEE/CVF International Conference on Computer Vision}, pages 1735--1744, 2019.

\bibitem[Zhou et~al.(2022)Zhou, Wang, Zhang, Sun, Zhang, Birchfield, Guo, Kong, Wang, and Zhong]{zhou2022audio}
Jinxing Zhou, Jianyuan Wang, Jiayi Zhang, Weixuan Sun, Jing Zhang, Stan Birchfield, Dan Guo, Lingpeng Kong, Meng Wang, and Yiran Zhong.
\newblock Audio--visual segmentation.
\newblock In \emph{European Conference on Computer Vision}, pages 386--403. Springer, 2022.

\bibitem[Zhou et~al.(2018)Zhou, Wang, Fang, Bui, and Berg]{zhou2018visual}
Yipin Zhou, Zhaowen Wang, Chen Fang, Trung Bui, and Tamara~L Berg.
\newblock Visual to sound: Generating natural sound for videos in the wild.
\newblock In \emph{Proceedings of the IEEE conference on computer vision and pattern recognition}, pages 3550--3558, 2018.

\bibitem[Zhu et~al.(2022)Zhu, Olszewski, Wu, Achlioptas, Chai, Yan, and Tulyakov]{zhu2022quantized}
Ye Zhu, Kyle Olszewski, Yu Wu, Panos Achlioptas, Menglei Chai, Yan Yan, and Sergey Tulyakov.
\newblock Quantized gan for complex music generation from dance videos.
\newblock In \emph{European Conference on Computer Vision}, pages 182--199. Springer, 2022.

\end{thebibliography}
}

\end{document}